\pdfoutput=1
\documentclass[11pt]{article}

\usepackage{authblk}
\usepackage[final]{acl}
\usepackage{times}
\usepackage{latexsym}
\usepackage{graphicx} 
\usepackage{float} %
\usepackage{subfigure} %

\usepackage{times}
\usepackage{graphicx}
\usepackage{url}
\usepackage{amsmath}
\usepackage{breqn}
\usepackage{makecell}
\usepackage{tabularx}
\usepackage{comment}
\usepackage{bm}
\usepackage{multirow}
\usepackage{microtype}
\usepackage{microtype}
\usepackage[utf8]{inputenc}

\usepackage{amssymb}
\usepackage{arydshln}
\usepackage{pgfplots}

\newcommand*{\affaddr}[1]{#1}
\newcommand*{\affmark}[1][*]{\textsuperscript{#1}}

\newcommand{\InitialErrorDistributionsBarChart}{

\begin{figure}
\begin{tikzpicture}[scale=.4]
    \begin{axis}[
        xbar, xmin=0,
        xmax=35,
        legend style={at={(.95,0.2)},nodes={scale=1.5}},
        height=14cm,
        enlarge y limits=.1,
        xlabel={},
        symbolic y coords={Modality,Tense,Object,Negation,Circumstance,Wrong Reference,Redundant Information,Missing Information,Total},
        ytick=data,
        nodes near coords, nodes near coords align={horizontal},
    ]
    
        \addplot[fill=green] coordinates {(28,Missing Information) (0,Redundant Information) (20,Wrong Reference) (16,Circumstance) (8,Negation) (8,Object) (8,Tense) (12,Modality)};
        
        \addplot[fill=blue] coordinates {(33,Missing Information) (22,Redundant Information) (12,Wrong Reference) (4,Circumstance) (6,Negation) (12,Object) (2,Tense) (8,Modality)};
        
        \addplot[fill=red] coordinates {(24,Missing Information) (3,Redundant Information) (16,Wrong Reference) (21,Circumstance) (18,Negation) (5,Object) (0,Tense) (13,Modality)};
        
        \legend{S-BART,D-HGN,BART}
    \end{axis}
\end{tikzpicture}
\caption{Percentage of error types in each model during preliminary human evaluation of 19 SAMSum dialogues.}
\end{figure}

}

\newcommand{\Dataset}{
\begin{table}
\small
\linespread{1.2}
    \centering
    \small
    \begin{tabular}{l|c|c|c|c}
    \hline
         & \textit{\textbf{Dialogue}} & \textit{\textbf{Speakers}} & \textit{\textbf{Turns}} &  \textit{\textbf{Length}}\\
    \hline
               \multicolumn{5}{c}{\it SAMSum} \\
    \hline

          Train & 14732 & 2.40 & 11.17 & 23.44 \\
          Validation & 818 & 2.39 & 10.83 & 23.42 \\
          Test & 819 & 2.36 & 11.25 & 23.12 \\
              \hline

           \multicolumn{5}{c}{\it AMI} \\
              \hline

          All  & 137 & 4 & 289 & 322\\
    \hline
    \end{tabular}
    \caption{Details about SAMSum and AMI.}
    \label{tab:dataset}
\vspace{-0.2cm}
\end{table}
}

\newcommand{\ErrorDistributionsAMI}{
\begin{table*}[ht]
\small
\centering
\begin{tabular}{lccccccc}
\hline

\textbf{Error Type} & {\textbf{BART}} &  {\textbf{BART-ConFiT}} & {\textbf{Pegasus}} &  {\textbf{Pegasus-ConFiT}} & {\textbf{T5}} &  {\textbf{T5-ConFiT}}  \\ \hline
Missing Information                 &         90\%                       &      85\%              &  80\%                              &       70\%               &      80\%                          &  85\%                      \\
Redundant Information                  &               10\%                 &     15\%               &    60\%                            &          25\%          &  0\%                              &            25\%                   \\
Wrong Reference               &   35\%                             &    30\%                &                  35\%              &   30\%                      &       50\%                         &            50\%                  \\
Circumstance &       35\%     &    35\%                            &        30\%            &                         30\%       &   40\%                                     &      35\%      \\
Negation                &      20\%                          &     15\%               &                     5\%           &    15\%                     &      25\%                          &      0\%                             \\
Object            &   45\%                             &         40\%           & 45\%                               &   25\%                        &             55\%                   &             55\%            \\
Tense &      10\%      &    10\%                            &      0\%              &   5\%                             &      10\%                                  &     10\%       \\
Modality       &    10\%                            &        15\%            &    5\%                            &          5\%                      &             20\%                   &   10\%             \\ \hline                            
\end{tabular} \caption{Percentage of autogenerated summaries containing each error type, according to our human evaluation of model outputs from 20 AMI dialogues. Note that a single summary can contain multiple error types, so they do not add up to 100\%.} \label{tab:distami}
\end{table*}}

\newcommand{\HumanFaithfulnessScores}{
\begin{table}[ht!]
\small
\linespread{1.2}
    \centering
    \small
    \begin{tabular}{l c c}
\hline \hline
\textbf{Faithfulness Score}          & \textbf{SAMSum} & \textbf{AMI}   \\ \hline \hline
BART           & 5.540  & 4.850 \\
BART-ConFiT    & \textbf{7.250}  & \textbf{5.600} \\ \hline
Pegasus        & 6.260  & 5.250 \\
Pegasus-ConFiT & \textbf{6.770}  & \textbf{5.895} \\ \hline
T5             & 5.422  & 4.150 \\
T5-ConFiT      & \textbf{6.920}  & \textbf{4.950} \\ \hline   \hline
\end{tabular} \caption{Average faithfulness score (on a scale of 1-10) given to each model by human evaluators on 100 SAMSum and 20 AMI dialogues. Highest scores for each dataset have been bolded.}
    \label{tab:humanfactuality}
\vspace{-0.2cm}
\end{table}
}

\newcommand{\BARTScores}{
\begin{table}[ht!]
\small
\linespread{1.2}
    \centering
    \small
\begin{tabular}{l c c}
\hline \hline
\textbf{BARTScore}          & \textbf{SAMSum} & \textbf{AMI}   \\ \hline\hline
BART           & -1.613  & \textbf{-3.644} \\
BART-ConFiT    & \textbf{-1.468}  & -3.669 \\ \hline
Pegasus        & -1.615  & \textbf{-2.967} \\
Pegasus-ConFiT & \textbf{-1.608}  & -3.369 \\ \hline
T5             & -1.993  & \textbf{-3.406} \\
T5-ConFiT      & \textbf{-1.677}  & -3.798 \\ \hline\hline
\end{tabular} \caption{Average BARTScore for each model on 100 SAMSum and 20 AMI dialogues. Highest scores for each dataset have been bolded.} \label{tab:bartscores}
\vspace{-0.2cm}
\end{table}
}

\usepackage[T1]{fontenc}

\newcommand{\CalD}{\mathcal{D}}

\usepackage{xspace}
\newcommand{\sys}{{\textsc{All}}\xspace}

\newcommand{\baby}{\textsc{ConFiT}\xspace}

\title{\baby: Toward Faithful Dialogue Summarization with\\Linguistically-Informed Contrastive Fine-tuning}

\author{\vspace{-0.4cm}

\textbf{Xiangru Tang}\affmark[$\dagger$] 
\quad
\textbf{Arjun Nair}\affmark[$\dagger$] 
\quad \textbf{Borui Wang}\affmark[$\dagger$]  
\quad \textbf{Bingyao Wang}\affmark[$\dagger$]  
  \quad \textbf{Jai Desai}\affmark[$\dagger$]\\
  \quad \textbf{Aaron Wade}\affmark[$\dagger$]
  {\bf 
  \enskip \textbf{Haoran Li}\affmark[$\P$]
  \enskip \textbf{Asli Celikyilmaz}\affmark[$\P$]
  \enskip \textbf{Yashar Mehdad}\affmark[$\P$]
  \enskip \textbf{Dragomir Radev}\affmark[$\dagger$]
  }
 \\
\affaddr{\affmark[$\dagger$] Yale University} 

  \affaddr{\affmark[$\P$] Meta AI} \\
{\tt{\{xiangru.tang, arjun.nair, borui.wang, dragomir.radev\}@yale.edu}} \\ 
{\tt{\{{aimeeli, aslic, mehdad\}@fb.com}}} \\

}

\begin{document}
\maketitle
\begin{abstract}
Factual inconsistencies in generated summaries severely limit the practical applications of abstractive dialogue summarization. Although significant progress has been achieved by using pre-trained neural language models, substantial amounts of hallucinated content are found during the human evaluation. 
In this work, we first devised a typology of factual errors to better understand the types of hallucinations generated by current models and conducted human evaluation on popular dialog summarization dataset. 
We further propose a training strategy that improves the factual consistency and overall quality of summaries via a novel contrastive fine-tuning, called \baby.
To tackle top factual errors from our annotation, we introduce additional contrastive loss with carefully designed hard negative samples and self-supervised dialogue-specific loss to capture the key information between speakers.
We show that our model significantly reduces all kinds of factual errors on both SAMSum dialogue summarization and AMI meeting summarization. On both datasets, we achieve significant improvements over state-of-the-art baselines using both automatic metrics, ROUGE and BARTScore, and human evaluation.

\end{abstract}

\section{Introduction}


Text summarization is used to generate a concise and accurate summary of a long text while focusing on the sections that convey the most useful information \cite{gurevych-strube-2004-semantic}. In recent years, the resurgence of dialogue summarization has attracted significant research attentions \cite{mccowan2005ami, gliwa-etal-2019-samsum, koay-etal-2020-domain, zhang-etal-2021-emailsum, zhong-etal-2021-qmsum, zhu2021mediasum, chen2021summscreen, li2021hierarchical, chen-etal-2021-dialogsum, fabbri-etal-2021-convosumm, chen2021dialogsum}. The goal of dialogue summarization is to condense the conversational input into brief sentences version but cover salient information \cite{mccowan2005ami, yuan2020abstractive}. 
Significant progress has been made recently on abstractive dialogue summarization with various pre-trained models. However, such pre-trained models are susceptible to generating hallucinate content that is not supported by the source documents \cite{cao2018faithful, maynez-etal-2020-faithfulness, kryscinski-etal-2020-evaluating}. 
To tackle the issue of factual inconsistency in dialogue summarization, recent works correctly encode the names of speakers \cite{zhu-etal-2020-hierarchical}, explicitly incorporate coreference information \cite{liu-etal-2021-coreference}, and order the personal named entities \cite{liu2021controllable}. But it is still challenging to improve the quality of summaries generated by different models and decrease the hallucination at the same time.

\begin{figure*}[h] 
\centering 
\includegraphics[scale=0.41]{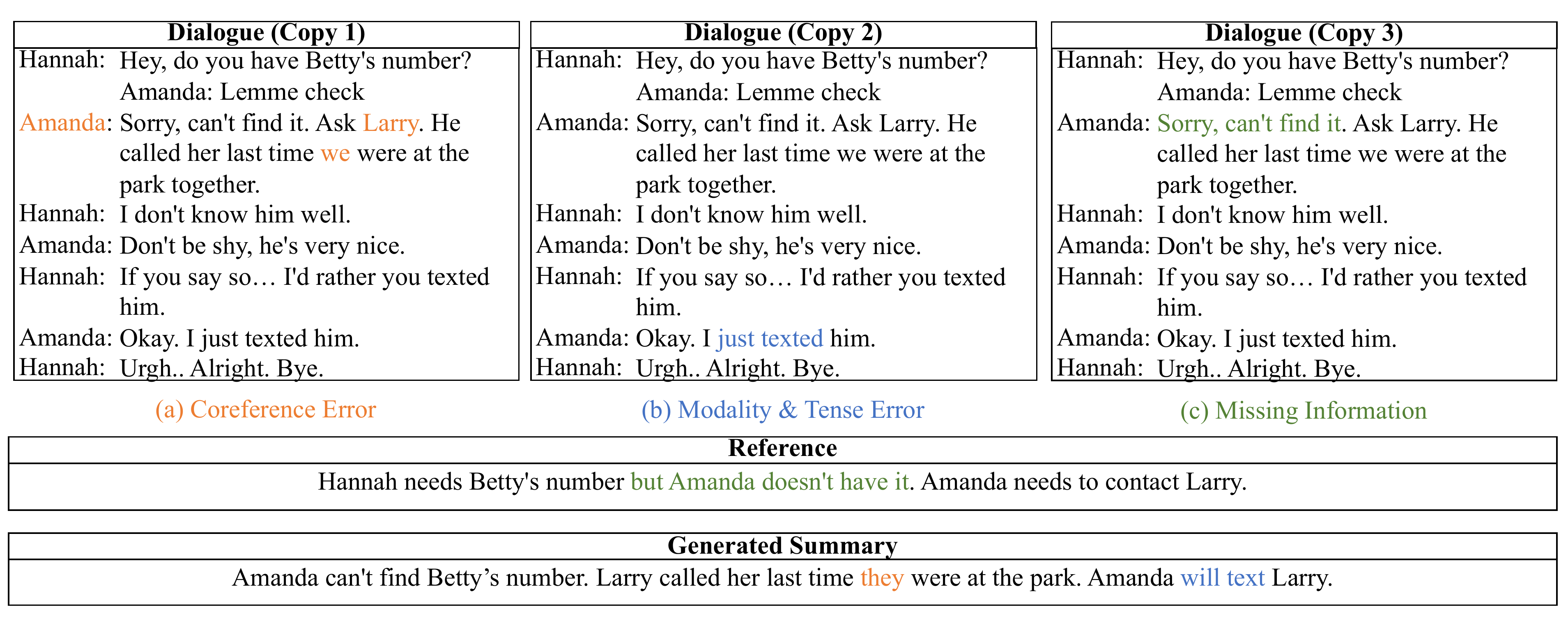} 
\caption{Sample summary of a SAMSum dialogue \cite{gliwa-etal-2019-samsum}. The summary is generated by BART \cite{lewis-etal-2020-bart}. Errors are highlighted.} 
\label{Fig.main1}
\end{figure*}



To better understand the types of hallucinations generated by the pre-trained models, we devised a linguistically motivated taxonomy of factual errors for dialogue summarization, instead of simply classifying the summary as faithful or not.
Based on our typology, we defined an annotation protocol for factuality evaluation of dialogue summarization. We then conducted a human evaluation of several pre-trained abstractive summarizers, including BART \cite{lewis-etal-2020-bart}, Pegasus \cite{pegasus}, and T5 \cite{2020t5}, aiming at identifying the proportion of different types of factual errors and studying the weaknesses of the pre-trained models. Our typology and annotation helps us gain deeper insights into the causes of factual inconsistency.
Unlike news summarization \cite{pagnoni-etal-2021-understanding}, we found that the challenges posed by dialogue summarization are more related to dialogue flow modeling, informal interactions between speakers, and complex coreference resolution. Figure \ref{Fig.main1} shows a dialogue-summary pair with three specific errors.

In order to tackle the top factual errors produced by existing models, we propose to replace the most commonly used fine-tuning with a linguistically-informed contrastive fine-tuning approach.
For example, the reason for producing wrong reference errors is that models cannot understand the role in the dialogue, which goes beyond the events.
Our goal is to drive the model to pay attention to the grounds of specific errors during the fine-tuning, and learn how to reduce the generation of such errors.
To be more specific, \baby learns to distinguish whether there are factual errors in the summaries and capture the key information in the dialogue content, such as numbers and person names.
Experiments on SAMSum \cite{gliwa-etal-2019-samsum} and AMI \cite{mccowan2005ami} show the generalizability of \baby when it is applied to different pre-trained models and datasets. 
Furthermore, we employ both automatic evaluation and human evaluation on faithfulness and show that \baby significantly reduces all different factual errors and generates summaries that are more factually consistent. Moreover, we analytically find that optimizing the contrastive fine-tuning is quite beneficial for improving the robustness of models, which brings further benefits.

Our contributions are as follows:

\begin{itemize}
\item We introduce the first typology of factual errors for dialogue summarization and use it to conduct comprehensive annotation and focused analysis.
\item Targeting different categories of factual errors in the annotations, we reduce occurrence of such errors generated by various pre-trained models with a novel linguistically-informed contrastive fine-tuning \baby approach.
\item We validate our method on a widely used dialogue summarization corpus, SAMSum, and extend it to  a meeting summarization corpus AMI. 
Evaluations of output summaries on automatic metrics like ROUGE, BARTScore as well as human evaluations show that \baby outperforms baseline pre-trained models.

\end{itemize}


\section{New Taxonomy of Factuality Errors for Abstractive Dialogue Summarization}\label{sec:taxonomy}

In order to gain deeper insights into the types of factuality errors introduced by different abstractive dialogue summarization systems, we proposed a new taxonomy of factuality errors for abstractive dialogue summarization based on our empirical experiments and annotations of the performance of a set of representative baseline summarization models on the SAMSum dataset, which is a widely-used large-scale dialogue summarization dataset of chat message dialogues in English (see Section \ref{sec:dataset}). Specifically, we generate summaries of SAMSum dialogues using state-of-the-art abstractive dialogue summarization models, including models fine-tuned based on T5 \cite{2020t5}, Pegasus \cite{pegasus}, BART \cite{lewis-etal-2020-bart}, D-HGN \cite{xiachong-etal-2021-incorporating}, and S-BART \cite{chen-yang-2021-structure}. We then manually annotate all different types of errors in these generated summaries that are inconsistent with the source dialogue, compute detailed statistics of all these factuality errors, and then classify them into different categories. Based on our annotation and analysis, we propose a new taxonomy of errors with the majority focusing on factuality error, which includes the following 8 error types:

\noindent\textbf{Category 1 - Missing Information:} The content of the generated summary is incomplete compared to the reference.

\textbf{Example:}
\begin{quote}
\indent [Reference Summary] \textit{Williams invites Ms. Blair for a coffee. They will go to her favourite coffee place near the square in a side alley at 2 p.m.} 

\indent [Model-Generated Summary] \textit{Ms. Blair is going to a coffee place near the square in a side alley.} 
\end{quote}

\noindent\textbf{Category 2 - Redundant Information:} There is redundant content in the generated summary compared to the reference.

\textbf{Example:}
\begin{quote}
\indent [Reference Summary] \textit{Paula helped Charlotte with correct pronunciation of "Natal Lily."} 

\indent [Model-Generated Summary] \textit{Charlotte asks Paula how to pronounce the name of the plant "Natal Lily." Paula confirms that the stress on the second syllable is 2nd.} 
\end{quote}

\noindent\textbf{Category 3 - Circumstantial Error:} Circumstantial information (e.g., date, time, location) about the predicate doesn't match the reference.

\textbf{Example:}
\begin{quote}
\indent [Reference Summary] \textit{The USA was founded in 1776.} 

\indent [Model-Generated Summary] \textit{The USA was founded in 1767.} 
\end{quote}

\noindent\textbf{Category 4 - Wrong Reference Error:} A pronoun is with an incorrect or nonexistent antecedent, or a personal named entity in the generated summary is in the place of a different personal entity in the reference.

\textbf{Example:}
\begin{quote}
\indent [Reference Summary] \textit{Mohit asked Darlene about the test.} 

\indent [Model-Generated Summary] \textit{Darlene asked Mohit about the test.} 
\end{quote}

\noindent\textbf{Category 5 - Negation Error:} This encompasses factual errors resulting from missing or erroneous negation in the generated summary compared to the reference.

\textbf{Example:}
\begin{quote}
\indent [Reference Summary] \textit{Justin likes books.} 

\indent [Model-Generated Summary] \textit{Justin does not like books.} 
\end{quote}

\noindent\textbf{Category 6 - Object Error:} This covers factual errors resulting from incorrect direct or indirect objects (for non-personal entities only; errors of this nature involving personal entities are designated as Wrong Reference Errors).

\textbf{Example:}
\begin{quote}
\indent [Reference Summary] \textit{Tara raised her glass.} 

\indent [Model-Generated Summary] \textit{Tara raised her spoon.} 
\end{quote}

\noindent\textbf{Category 7 - Tense Error:} This encompasses factual errors resulting from discrepancies in grammatical tense between the generated summary and the reference.

\textbf{Example:}
\begin{quote}
\indent [Reference Summary] \textit{The children will go to the library.} 

\indent [Model-Generated Summary] \textit{The children went to the library.} 
\end{quote}

\noindent\textbf{Category 8 - Modality Error:} This includes factual errors resulting from modal discrepancies, such getting words like "may", "should", "could" wrong, between the generated summary and the reference.

\textbf{Example:}
\begin{quote}
\indent [Reference Summary] \textit{School may be cancelled today.} 

\indent [Model-Generated Summary] \textit{School is cancelled today.} 
\end{quote}

\subsection{Annotation and Analysis}

\InitialErrorDistributionsBarChart

Using our proposed taxonomy of factuality errors, we compute the proportion of each type of factuality errors across different summarization models. 
We then investigate the model generation behavior that is indicative of errors, which guides the design of our proposed model.

We performed a human evaluation of four model outputs from 19 SAMSum dialogues in order to identify the limitations of abstractive summarization models in dialogue summarization tasks. The four models used in this human evaluation are two BART models with different random seeds (ROUGE-L 48 and 49) \cite{lewis-etal-2020-bart}, D-HGN (ROUGE-L 40) \cite{xiachong-etal-2021-incorporating}, and S-BART (ROUGE-L 48 \cite{chen-yang-2021-structure}). BART and S-BART are pre-trained models (PLM), and D-HGN is trained from scratch. Since we are focusing on the dialogue domain, most of the factual errors in the model summaries are related to coreference, anaphora, and other dialogue-specific characteristics. In fact, approximately 45\% of all errors fall into the categories of Missing Information and Wrong Reference. The distribution of these errors throughout these pre-existing models informs the limitations of each model. Our proposed \baby model targets the top errors generated by the current state-of-the-art models to reduce factual inconsistency.

\section{\baby Model}

Standard fine-tuning parameterizes the probability $p_\alpha$ of the generator on a task-specific labeled dataset by maximizing cross-entropy loss. 
\vspace{-5mm}

\begin{align}
\centering
    \mathcal{L} = - \sum \log P(\tilde{t}_l|t_{<l}, \mathbf{D}) 
\end{align}

However, the cross-entropy loss has several shortcomings that can lead to factual inconsistency in dialogue summarization due to its sub-optimal generalization and instability. We propose a more efficient fine-tuning method \baby for factual consistency driven by the intuition that good generalization requires capturing the similarity in one class and contrasting them in other classes.
In \baby, we introduce two additional losses: contrastive loss and self-supervised loss. We use two weights, actually which is coefficients, to adjust the ratio of $L_{con}$ and $L_{self}$ in the total loss of \baby.  The final training objective $\mathcal{J}(\theta)$ of the proposed framework is as follows:

\begin{equation}
\centering
\setlength{\abovedisplayskip}{3pt}
\setlength{\belowdisplayskip}{3pt}
    \mathcal{J}(\theta) = \mathcal{L} +  \alpha \mathcal{L}_{\text{con}} +\beta \mathcal{L}_{\text{self}} 
\end{equation}

\begin{figure}[h] 
\centering 
\includegraphics[scale=0.38]{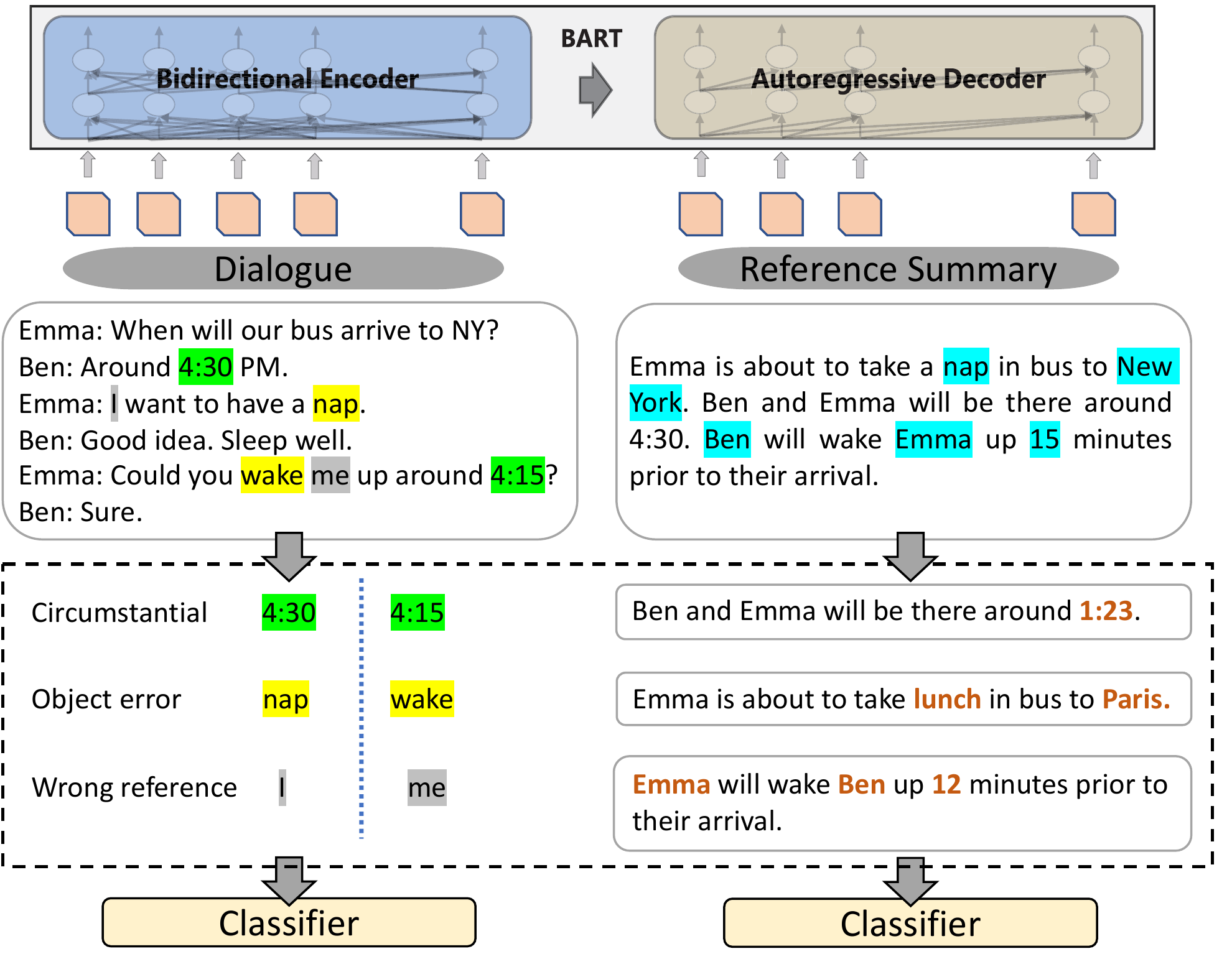} 
\caption{A demonstration of our model.} 
\label{Fig.model1}
\end{figure}

Our linguistically-informed typology and annotation help us gain deeper insights into the causes of different factual errors. To help our models generate more faithful summaries, the proposed \baby learns to concentrate on the essential elements of dialogue and capture the dynamic role information as illustrated in Figure \ref{Fig.model1}.

\vspace{-3mm}


\subsection{Contrastive Loss}


In order to reduce the occurrence of factual errors, we propose a contrastive loss that uses the following negative sample generation techniques to target each error type in our proposed taxonomy:

\begin{itemize}
    \item Swap the nouns in the reference summary with each other randomly. This aims to reduce wrong reference and object errors by providing negative samples.
    \item Swap the verbs in the reference summary with each other randomly. This aims to the model reduce circumstance (and, to a lesser extent, tense and modality) errors.
    \item Mask numbers and years in the dialogue and then pass it into the model to generate a negative sample summary. This aims to reduce circumstance errors.
    \item Randomly delete 30\% of the sentences in the dialogue and then pass it into the model to generate a negative sample summary. This aims to reduce missing information errors.
    \item Mask-and-fill coreferent entities with BART in the dialogue and then pass it into the model to generate a negative sample summary. This aims to reduce wrong reference errors.
\end{itemize}




\begin{table*}[t]
\small
\centering
        \begin{tabular}{l||ccc||ccc}
            \hline            \hline
            &  \multicolumn{3}{c||}{\textbf{AMI}} & \multicolumn{3}{c}{\textbf{SAMSum}} \\

            \textbf{Model} & \textbf{R-1} & \textbf{R-2} & \textbf{R-L} & \textbf{R-1} & \textbf{R-2} & \textbf{R-L}  \\
            \hline
            \hline
            \multicolumn{7}{c}{\it Extractive and Abstractive Models} \\
            \hline   
            TextRank \cite{textrank} &35.19* &6.13* &15.70* &29.27* &8.02* &28.78*  \\
            Fast Abs RL \cite{chen-bansal-2018-fast}   &38.76 & 15.13 & 35.18          & 40.96 & 17.18 & 39.05 \\
            PGN \cite{pgn} &48.34* &16.02* &23.49* & 40.08* & 15.28* & 36.63* \\
         PGN($\CalD_{\sys}$) \cite{acl2021-gpt-feng} &50.91* &17.75*  &24.59* & - & - & - \\
            \hline
            \hline
            \multicolumn{7}{c}{\it Pre-trained Models} \\
            \hline 
       T5 \cite{2020t5}              &42.16  &13.94 & 39.39    & 48.41 & 24.79 & 44.61 \\
Pegasus \cite{pegasus}   &46.02  &15.85 &43.73   &48.04 & 22.94 & 43.40     \\
       BART \cite{lewis-etal-2020-bart}  &47.92  & 16.00& 45.36  &51.74 & 26.46& 48.72  \\
       Multi-view BART \cite{chen-yang-2020-multi}&-&-&-& 49.52& 26.52& 48.29 \\
            \hline
            \hline
            \multicolumn{7}{c}{\it Ours} \\
            \hline
 T5-ConFiT   &47.18 &13.19 &43.55  &  52.13 & 27.12 & 47.62  \\
Pegasus-ConFiT &48.47  &17.61 &45.75   & 52.65 & 28.21 & 48.15
  \\
BART-ConFiT   &50.31  &17.29 & 47.98 &53.89 & 28.85 &  49.29    \\
            \hline            \hline
        \end{tabular}
\caption{Dialogue summarization ROUGE evaluation on the AMI \protect\cite{mccowan2005ami} and SAMSum \protect\cite{gliwa-etal-2019-samsum} datasets. We adopt some results reported from the literature \cite{feng2021survey} and implement the pre-trained models for a  fair comparison.
All results marked with an asterisk (*) are from \citet{acl2021-gpt-feng}.} 
\label{tab:testtable}
\end{table*}


Equation \ref{eq:con} demonstrates our contrastive loss function. During the fine-tuning, we have the positive samples, which is the reference summaries and another set of incorrect summaries, which is the negative samples.  The contrastive objectives are learning representations that are invariant to different views of positive pairs; while maximizing the distance between negative pairs \cite{gunel2020supervised}. Our goal is to maximize the likelihoods of the positive samples and minimize the likelihoods of the negative samples as well.
We use the following contrastive learning objective

\vspace{-0.5cm}
\begin{equation}
\centering
    \fontsize{10}{11}\selectfont
    \mathcal{L}_{con} = - \sum_{\substack{y_j \ne y_i}}
    \log \frac{\exp(\mathrm{cos} (\bm{c}_i, \bm{c}_j))}{\sum\limits_{\substack{y_k \ne y_i}} \exp(\mathrm{cos} (\bm{c}_i, \bm{c}_k))}
    \label{eq:con}
\end{equation}

where $y_i$ and $y_j$ are positive summary pairs generated by back translation technology and $y_k$ is from negative set of examples and $\bm{c}_i$ ,$\bm{c}_j$, $\bm{c}_k$ are their BART decoder representations.




\subsection{Self-supervised Loss}

One unique challenge in abstractive dialogue summarization is the use of first-person pronouns (such as "I" or "we") in speaker utterances, which the model has to correctly identify as being a reference to the speaker. This can lead to wrong reference errors in the summary, as the model cannot understand which participant is speaking and thus cannot accurately resolve first-person references. To address this problem, we design a self-supervised loss that aims to determine whether two tokens belong to the same speaker. Based on these findings, we design a self-supervised loss to enable \baby to capture the dynamic roles in the dialogue.

After the BART encoder, the input dialogue is encoded into hidden vectors $C$. Here, we first randomly select $k$ pairs of two tokens $t_m$ and $t_n$ from the input dialogue, with labels $s_m$ and $s_n$ denoting which speaker they are coming from. We also do the same for utterances. Given the concatenation of the encoder representation of dialogue, $t_m$ and $t_n$, we use the following loss function to classify whether the two tokens or two utterances are from the same speaker.

\vspace{-6mm}

\begin{equation}
    \fontsize{10}{11}\selectfont
\mathcal{L}_{self}= - \sum_{m=1}^{k}\sum_{n=1}^{k}\log{P(s_m = s_n | t_m,t_n,C)}
\end{equation}

This supplementary loss function helps \baby keep track of speaker information, thus improving the faithfulness of its summaries for dialogues that contain several first-person references.

\section{Experiments}

\subsection{Dataset} \label{sec:dataset}

\Dataset
We evaluate our new model on the popular SAMSum dialogue summarization dataset. Then, we extend our model to meeting summarization with the AMI Meeting Corpus. SAMSum \cite{gliwa-etal-2019-samsum} is a recently proposed large-scale dialogue summarization dataset consisting of 16,369 chat message dialogues in English written by linguists, and each message dialogue is annotated with a multi-sentence summary written by language experts. 75\% of the dialogues in the SAMSum dataset \cite{gliwa-etal-2019-samsum} are between two interlocutors, and the other 25\% are among three or more interlocutors. The AMI Corpus is another well-known dialogue summarization dataset consisting of 137 multiparty meeting transcripts extracted from 100 hours of meeting recordings. Each meeting transcript in the dataset is also annotated with a generic abstractive summary. We use these two representative dialogue summarization datasets to empirically test our new model’s abstractive summarization performance in the settings of both short conversation-style dialogues and long meeting-style dialogues. See Table \ref{tab:dataset} for detailed statistics of the two datasets.
\vspace{-0.2cm}

\subsection{Experiment Settings}

In our experiment using SAMSum, we trained BART for 3 epochs with a learning rate of $1e-05$, Pegasus for 20 epochs with a learning rate of $1e-04$, and T5 for 20 epochs with a learning rate of $1e-05$. In our experiment using AMI, we trained BART for 6,000 steps with a learning rate of $1e-05$, Pegasus for 24,000 steps with a learning rate of $1e-05$, and T5 for 20,000 steps with a learning rate of $1e-05$. 


\begin{table*}[ht]
\small
\centering
\begin{tabular}{lccccccc}
\hline

\textbf{Error Type} & {\textbf{BART}} &  {\textbf{BART-ConFiT}} & {\textbf{Pegasus}} &  {\textbf{Pegasus-ConFiT}} & {\textbf{T5}} &  {\textbf{T5-ConFiT}}  \\ \hline
Missing Information                 &         55\%                       &      44\%              &  56\%                              &       50\%               &      63\%                          &  48\%                      \\
Redundant Information                  &               12\%                 &     7\%               &    7\%                            &          4\%          &  7\%                              &            4\%                   \\
Wrong Reference               &   37\%                             &    17\%                &                  25\%              &   18\%                      &       46\%                         &            13\%                  \\
Circumstance &       14\%     &    8\%                            &        16\%            &                         10\%       &   8\%                                     &      9\%      \\
Negation                &      4\%                          &     1\%               &                     7\%           &    2\%                     &      1\%                          &      1\%                             \\
Object            &   10\%                             &         6\%           & 4\%                               &   7\%                        &             2\%                   &             7\%            \\
Tense &      2\%      &    1\%                            &      3\%              &   1\%                             &      2\%                                  &     2\%       \\
Modality       &    6\%                            &        1\%            &    3\%                            &          5\%                      &             5\%                   &    8\%             \\ \hline                            
\end{tabular} \caption{Percentage of autogenerated summaries containing each error type, according to our human evaluation of model outputs from 100 SAMSum dialogues. Note that a single summary can contain multiple error types, so they do not add up to 100\%.} \label{tab:distsamsum}
\end{table*}

\vspace{-0.1cm}

\ErrorDistributionsAMI

\subsection{Evaluation Metrics}

To evaluate our model, we use three metrics: 

\textbf{ROUGE} \citep{lin-2004-rouge}: ROUGE measures N-gram overlap between the reference and the automatically generated summaries.

\textbf{BARTScore} \citep{yuan2021bartscore}: Because ROUGE scores only measure token overlap, other automated metrics \cite{rebuffel2021data,kryscinski-etal-2020-evaluating,wang-etal-2020-asking, scialom2021questeval} have been proposed to evaluate faithfulness more precisely. BARTScore is a transformer-based measure that scores a dialogue and the corresponding automatically generated summary and has been shown to be strongly correlated with human evaluations of faithfulness \citep{yuan2021bartscore}.

\textbf{Human Evaluation}: Finally, we conduct human evaluations on 100 SAMSum \cite{gliwa-etal-2019-samsum} and 20 AMI \cite{mccowan2005ami} dialogues.  \citet{tang2021investigating} found that Likert scales are a more consistent measure of factuality for abstractive dialogue summarization than Best-Worst Scaling. We have human evaluators directly rate the summaries on a scale from 1 to 10 corresponding to their faithfulness. In addition, using the error taxonomy proposed in Section \ref{sec:taxonomy}, we have them mark whether each error type appeared in the given summary. We do this in a blinded fashion, so that the annotators do not see the corresponding model of the summary. Additionally, in order to prevent model information from leaking to the annotators, we randomly shuffle outputs within each dialogue before assigning them to annotators.

\vspace{-0.05cm}
\section{Results}

\HumanFaithfulnessScores

\BARTScores

Table \ref{tab:testtable} shows the ROUGE scores of our models, the baseline models they were fine-tuned from, and a number of other abstractive summarization models on the SAMSum and AMI datasets. Tables \ref{tab:humanfactuality} and \ref{tab:bartscores} show the average human faithfulness and BART scores respectively for each model's outputs on 100 SAMSum and 20 AMI dialogues.

We observe that for all three pretrained models \baby significantly beat baselines on ROUGE-1, ROUGE-L, and human faithfulness score for both datasets. For BARTScore, we note that, while performance increases on SAMSum for all models, it decreases on AMI. However, given the fact that human evaluators rated the outputs of all three \baby models as more faithful than those of their corresponding baselines on both datasets, the decreases in BARTScore on AMI can likely be attributed to the imperfection of automated metrics at capturing faithfulness in text.

\subsection{Error Analysis}

Tables \ref{tab:distsamsum} and \ref{tab:distami} show the percentage of summaries that were labeled with each error type in our taxonomy of factual errors (discussed in Section \ref{sec:taxonomy}.) for both the baseline and \baby models on the SAMSum and AMI datasets respectively. 

We observe that on SAMSum, our fine-tuning method greatly reduces missing information, redundant information, wrong reference, and circumstance errors for all models. The largest reduction is on the "wrong reference" error type (20\%, 7\%, and 33\% for BART, Pegasus, and T5 respectively), likely owing to the self-supervised loss function introduced in Section 3.2 that was designed to help the model more effectively capture speaker information. 
For AMI, however, our fine-tuning method is not as consistent at reducing the frequency of each error type across models. It is possible that this is due to sample size (20 AMI dialogues vs. 100 SAMSum dialogues).

\subsection{Case Study}

\begin{figure*}[ht!] 
\centering 
\includegraphics[width=\textwidth]{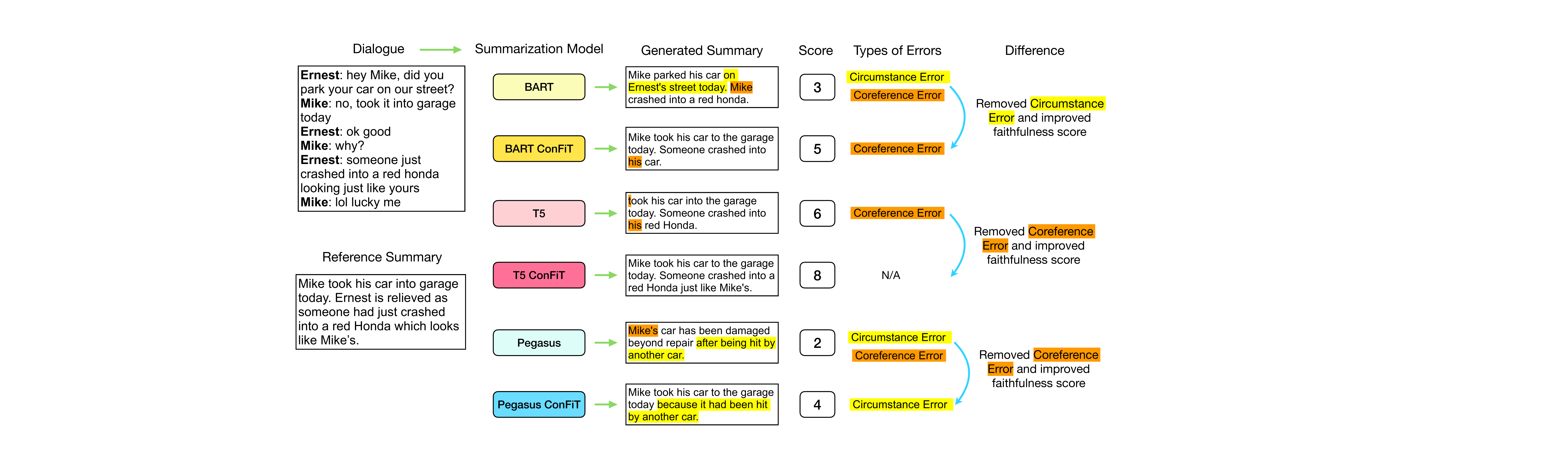} 
\caption{Model outputs for selected SAMSum dialogue, along with the corresponding reference summary, human factuality scores, and errors.} 
\label{fig:casestudy}
\end{figure*}


Figure \ref{fig:casestudy} shows the results of human annotation on the model outputs of a selected SAMSum dialogue. Note that all of the autogenerated summaries, both baseline and \baby, were marked as having missing information errors by the annotator, likely due to the omission of Ernest's relief upon hearing that the car that was crashed into did not belong to Mike. As a result, none of the models achieved a perfect factuality score on this dialogue; however, the scores for each \baby model were higher than those of their corresponding baselines.

It can be observed that while baseline BART outputs a summary with a circumstance error, mistakenly asserting that Mike parked his car on Ernest's street, the BART+\baby fixes this error, correctly asserting that Mike took his car to the garage today; as a result, the human annotator gave this summary a higher score than the predicted summary from baseline BART. Baseline T5 outputs a summary with two coreference errors; specifically, it contains a missing subject in the first sentence and incorrectly implies that the car that got crashed into belonged to Mike in the second sentence. The T5+\baby is able to fix both of these errors, adding \textit{"Mike"} to the beginning of the first sentence and changing \textit{"his red Honda"} to \textit{"a red Honda just like Mike's"} in the second sentence. Similarly, the output of baseline Pegasus contains a coreference error in the first sentence, implying that Mike owns the car that was crashed into while the output of Pegasus+\baby does not.






\section{Related Work}

Multi-party dialogues are especially challenging to summarize using automated models, given that they often contain pauses, false starts, reconfirmations, hesitations, and speaker interruptions \cite{sacks1978simplest, feng2021survey, chen2021simple}. Previous work in the field has addressed these challenges by incorporating semantic features, including keywords~\cite{zhu-etal-2020-hierarchical}, domain terminologies~\cite{koay-etal-2020-domain}, topics~\cite{zhao-etal-2020-improving, liu2021topicaware}, entailment knowledge~\cite{li-etal-2018-ensure}, and background knowledge~\cite{feng-etal-2021-language}. Other works exploit personal named entities~\cite{liu2021controllable} and coreference information~\cite{liu-etal-2021-coreference} to learning to distinguish complex coreferent relationships expressed through personal pronouns (including the first person "I") in the conversation \cite{lei2021hierarchical}.
Researchers have also explored conversational structure \cite{zhao2021improving}, utterance flow modelling \cite{chen2021dialogue}, syntactic structure \cite{lee2021says}, granularity control \cite{wu-etal-2021-controllable}, but they have not yet converged to a simple and practical solution.

Our proposed taxonomy of factual errors and annotations help us gain deeper insights into the causes of factual inconsistency in abstractive dialogue summarization outputs.

\section{Conclusion}

We presented \baby, a novel method to improve the faithfulness of abstractive dialogue summarization models via contrastive and self-supervised fine-tuning. By adapting the objective function during fine-tuning to incorporate a contrastive loss that learns to distinguish positives from examples with factual errors, and a self-supervised dialogue-specific loss that captures important dialogue information flow between multiple interlocutors, \baby can significantly improve the faithfulness of the abstractive summaries generated by transformer-based sequence-to-sequence language models, and reduce multiple categories of factuality errors in the abstractive summaries by large margins. In our experiment on SAMSum and AMI, we demonstrated that \baby achieves better empirical performance compared to the baseline models fine-tuned with the traditional cross-entropy loss, based on both automatic evaluation metrics and human evaluation. Our work provides new insights into improving the faithfulness of abstractive summarization systems using carefully designed novel objective functions for fine-tuning that captures important structures and features of the text to summarize.

\newpage
\section{Ethics Statement}

\paragraph{Human Evaluation} We recruited seven volunteer participants for our error annotation, requesting speakers of English. The internal annotators are Xiangru Tang, Arjun Nair, Borui Wang, Jai Desai, Aaron Wade, Anushka Nijhawan, and Dragomir Radev. These annotators are participating voluntarily. Our participants are free to opt out of the study at any point in time. We have written four scripts for use in the annotation process: (1) the first script generates an annotation spreadsheet and a key spreadsheet from the model outputs. The annotation spreadsheet does not contain the model names; however, it contains an id that can be used to recover the model name from the key spreadsheet. For ease of annotation, summaries from the same dialogue are grouped together; however, they are randomly shuffled within each dialogue so that the annotators cannot guess from the ordering as to which model is which. (2) The second script splits an annotation spreadsheet into multiple spreadsheets so that the work can be distributed amongst annotators. (3) The third one merges these spreadsheets back together after the annotation process is finished. (4) The last script recovers the model names from the key spreadsheet and inserts them into the annotation spreadsheet. Each evaluator is asked to examine whether there is an error and the full context (dialogue, generated summaries, and reference) and give a score on a scale of 1 to 10 for each of the criteria. We only consider faithfulness, instead of general quality. E.g. 1: very poor, 3: poor, 5: neutral; 7: good; 10: very good. We asked each internal annotator to evaluate 300 samples.

\paragraph{Other Ethical Issues} (1) We did not use any personally identifiable information in the experiments. (2) The goal of the project, improving the faithfulness of automatically generated summaries, is to make the output of the summarization system more reliable and minimize confusion for the readers of the summaries. (3) We used existing summarization datasets that do not contain any sensitive information and are unlikely to cause any harm to the annotators.


\newpage
\bibliography{custom}
\bibliographystyle{acl_natbib}

\end{document}